\documentclass[runningheads]{llncs}

\usepackage[mobile]{eccv}

\usepackage{eccvabbrv}

\usepackage{graphicx}
\usepackage{booktabs}
\usepackage{subcaption}
\usepackage[accsupp]{axessibility}  

\usepackage{multirow}
 \usepackage{bm} 

\usepackage[pagebackref,breaklinks,colorlinks]{hyperref} 

\usepackage{orcidlink}

\begin{document}

\title{CasSR: Activating Image Power for Real-World Image Super-Resolution} 

\titlerunning{Abbreviated paper title}

\author{Haolan Chen\inst{1,2} \and
Jinhua Hao\inst{2} \and
Kai Zhao\inst{2} \and Kun Yuan \inst{2} \and Ming Sun \inst{2} \and Chao Zhou \inst{2} \and Wei Hu \inst{1}}


\institute{Peking University, China, Beijing \and
Kuaishou Technology, China, Beijing \\
\email{\{chenhl99,forhuwei\}@pku.edu.cn}\\
\email{\{haojinhua,zhaokai05,sunming03,yuankun03,zhouchao\}@kuaishou.com}}

\maketitle

\begin{abstract}
The objective of image super-resolution is to generate clean and high-resolution images from degraded versions. Recent advancements in diffusion models have led to the emergence of various image super-resolution techniques that leverage pre-trained text-to-image (T2I) models. Nevertheless, due to the prevalent severe degradation in low-resolution images and the inherent characteristics of diffusion models, achieving high-fidelity image restoration remains challenging. Existing methods often exhibit issues including semantic loss, artifacts, and the introduction of spurious content not present in the original image.
To tackle this challenge, we propose \textbf{Cas}caded diffusion for Super-Resolution(\textbf{SR}), \textbf{CasSR}, a novel method designed to produce highly detailed and realistic images. 
In particular, we develop a image activation module that aims to optimize the extraction of information from low-resolution images. This module generates a preliminary reference image to facilitate the initial extraction of information and the mitigation of degradation. Furthermore, we propose a multiple attention mechanism to enhance the T2I model's capability in maximizing the restoration of the original image content. Through a comprehensive blend of qualitative and quantitative analyses, we substantiate the efficacy and superiority of our approach. The source code will be released upon acceptance of the paper.
\end{abstract}

\section{Introduction}
\label{sec:intro}
\begin{figure}[tb]
  \centering
  \includegraphics[height=5.5cm]{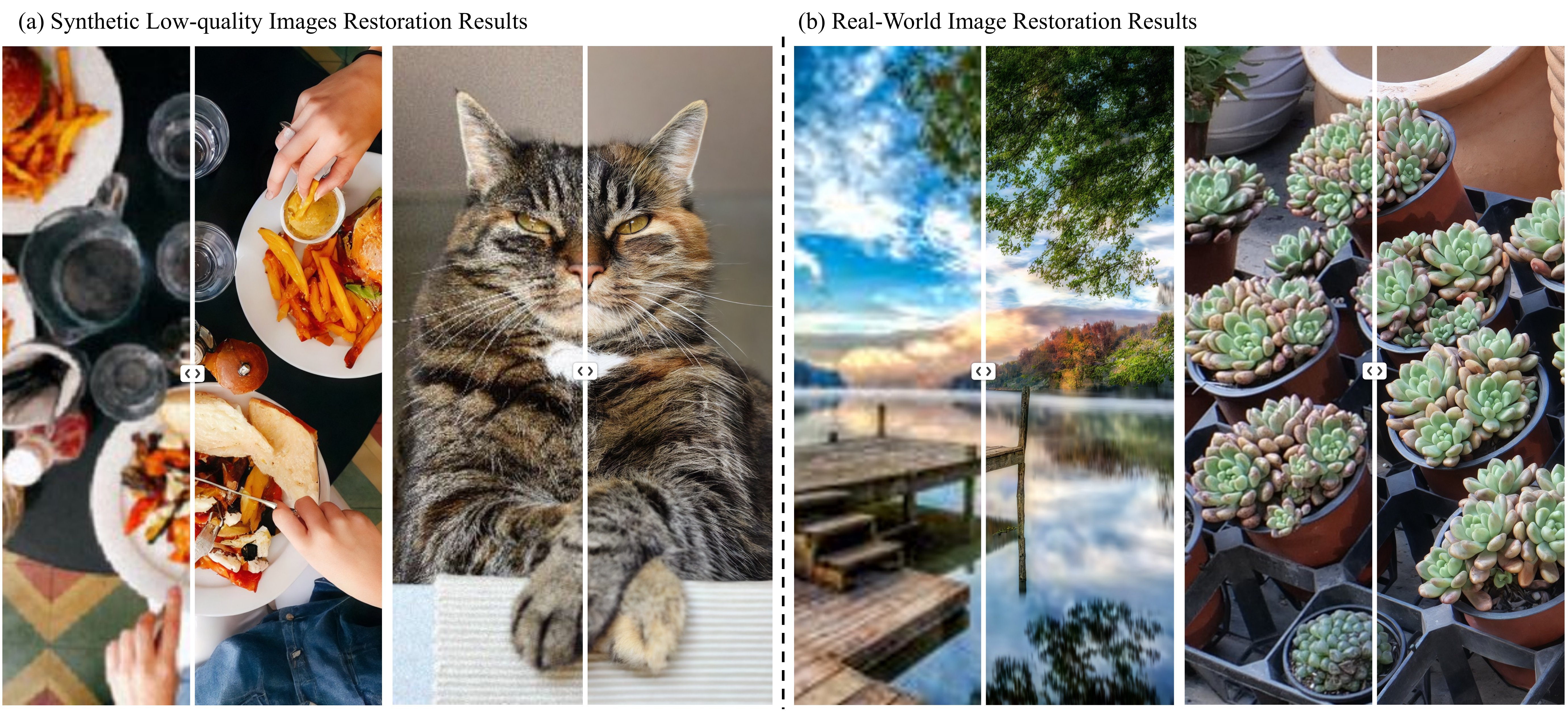}
  \caption{Our model has demonstrated strong image super-resolution capabilities. As shown above, CasSR has achieved outstanding results on both real-world and synthetic data.}
  \label{fig:example}
\end{figure}
Images inevitably suffer from degradation such as low resolution (LR), blur, and noise during the process of acquisition and transmission \cite{liu2023ada,yuan2023capturing}. Therefore, image super-resolution (ISR) has become a crucial task with broad applications in various fields such as medicine, autonomous driving and robotics \cite{chen2019camera, wan2020deep, umehara2018application, yue2016image}. 

The objective of ISR is to enhance the quality of images by reconstructing high-resolution (HR) versions from degraded input images. This problem is inherently challenging due to its ill-posed nature, leading to a surge in interest among researchers. Traditional methods \cite{chen2021pre, dong2014learning, dong2015image, yoon2015learning, zhang2018image, zhang2022efficient,wang2021realt}, however, have often relied on specific assumption about the degradation process. It is obvious that such assumptions are mostly impractical and may not hold in real-world scenarios. Moreover, these conventional approaches tend to produce over-smoothed results, thereby compromising the preservation of fine details in the images. 

To overcome these limitations, recent advancements in this field have attempted methods based on generative adversarial networks (GANs) \cite{goodfellow2020generative}. These models have demonstrated promising capabilities in addressing the real ISR problem, where the degradation factors are unknown, thus opening up new possibilities for achieving high-quality ISR in practical applications. However, they may not be sufficient for highly degraded inputs in complex scenarios \cite{liang2022details}. The trade-off in GAN-based approaches, prioritizing visual details over fidelity, can lead to the generation of visual artifacts that diminish overall image quality. Moreover, overcoming the substantial domain gap between training and testing data remains a challenging task in the realm of ISR.

Attributed to their excellent performance, Denoising Diffusion Probabilistic Models (DDPMs) \cite{ho2020denoising} have emerged as a focal point of interest in recent years. These models have consistently outperformed Generative Adversarial Network (GAN)-based methods across a spectrum of downstream tasks \cite{rombach2022high, blattmann2023align, kingma2021variational,kong2020diffwave, mittal2021symbolic,saharia2022image}, marking a significant milestone in the field. To tackle the intricate challenge of image degradation, particularly when the specific nature of degradation remains unidentified, the research community has increasingly leaned towards leveraging large-scale pre-trained text-to-image (T2I) models \cite{Rombach_2022_CVPR}. These models, trained on expansive datasets comprising over 5 billion image-text pairs \cite{schuhmann2022laion}, offer a robust foundation and abundant prior knowledge for addressing complex image restoration tasks.
One pioneering work, ControlNet \cite{zhang2023adding}, introduced an innovative approach to harness extra conditional information, such as edge maps and depth maps, to steer the generative capabilities of pre-trained models towards desirable outcomes. Inspired by this, a series of subsequent studies \cite{wang2023exploiting, yang2023pixel, lin2023diffbir, wu2023seesr, sun2023coser} are then proposed to generate high-quality images exploiting T2I models' strong prior. These efforts underscore the potential of T2I models in transcending traditional image restoration techniques by incorporating rich and contextual information.

However, the highly generative nature of T2I models presents a double-edged sword. While they possess the ability to fill in missing information and enhance image quality, they are also prone to deviating from the ground truth content \cite{wang2023exploiting, yang2023pixel}, sometimes introducing elements that were not present in the ground truth. This propensity raises concerns about the fidelity of the restored images, especially in complex scenarios featuring multiple objects and severe degradation. The challenge of preserving authenticity and accuracy, therefore, remains crucial in the development of these diffusion-based ISR models. To address this issue, some studies attempt to design more suitable text prompts \cite{wu2023seesr,yang2023pixel}, or aim to include more information in the semantic embedding \cite{wu2023seesr,sun2023coser}. On the contrary, We argue that the primary reason is that many methods only utilize the original LR image as the conditional input for the pre-trained T2I model, while failing to fully leverage the information in the input image and the prior knowledge of the diffusion model. Enhancing the quality of the input LR image is a more essential way to improve the output results.

In this paper, we propose a novel method, \textbf{Cas}caded diffusion for Super-Resolution(\textbf{SR}), \textbf{CasSR}, to solve aforementioned problems. The key idea of our method is to further exploit the information in the input image and minimize the impact of degradation in the input image on the generation process as much as possible.

Firstly, we cascade the diffusion model by adopting SCEdit \cite{jiang2023scedit} to attenuate the degradation of the input image before sending it to the U-Net. SCEdit is not specifically designed for ISR, but it is a light-weighted and effective model for controllable generation with pre-trained SD models, hence it significantly reduces extra computational complexity. With the initial recovery, more information can be mined for subsequent steps. To further refine the restoration process, we introduce a multiple attention module, moving beyond the conventional use of simple skip connections of ControlNet \cite{zhang2023adding}. The first stage output and the LR image are input into the multiple attention module, providing a more precise control over the generation process. This ensures that the restored images maintain their quality while displaying improved clarity and detail. Through extensive qualitative and quantitative experimental validation, we demonstrate the effectiveness of our approach in achieving high-quality image restoration.

The contribution of this paper can be summarized as follows:
\begin{enumerate}
    \item We propose a novel model CasSR for image super-resolution, aiming to generate high-quality images with strong fidelity. By adding control of the input image to the model, CasSR fully exploits the prior knowledge of diffusion models. This model is also flexible as various methods can be employed as the image activation module.
    \item We highlight the significance of image guidance over semantic information, showing that a simple enhancement of the input LR image can lead to significant improvements in the output. Thus, we offer a new perspective on research unrestricted by textual limitations.
    \item We introduce a novel multiple attention architecture to make full use of the information in the input image to generate results with high fidelity. Qualitative and quantitative experimental results across multiple benchmarks validate the effectiveness of our model.
\end{enumerate}

\section{Related Work}
\label{sec:rel}
\subsection{Deep-learning-based Image SR}
Image super-resolution aims at restoring an high resolution (HR) image from its corresponding low resolution (LR) input. Since the pioneer work SRCNN \cite{dong2014learning} is proposed, various methods \cite{chen2023activating, dong2016accelerating, ren2017image, dai2019second, liang2021swinir, lim2017enhanced,zhang2022efficient,zhang2018residual,xu2018dense} have attempted to employ deep neural networks for image super-resolution. However, they usually assume that the degradation process is simple, for example, bi-cubic down-sampling, blurring and noising. These methods may perform well on specific dataset with known degradation. When dealing with real-world super-resolution problem whose degradation is much more complex, they may often fail due to lack of generalization ability. \cite{gu2024networks, liu2022discovering}. Moreover, since their training objective are mainly pixel-wise metrics, e.g. PSNR and SSIM \cite{wang2004image}, they usally tend to generate over-smoothed results.

For improved generalization, recent studies have explored the use of generative adversarial networks (GANs) \cite{goodfellow2020generative}. Additionally, in order to more accurately replicate real-world low-resolution images, BSRGAN \cite{zhang2021designing} and Real-ESRGAN \cite{wang2021real} have introduced their own degradation processes. After that, a series of methods like SwinIR \cite{liang2021swinir} and HAT \cite{chen2023activating} exploits the generative ability of transformer to reconstruct the HR image. Thanks to improved model generation capabilities and better training data, these methods have shown promising results \cite{liang2022details}. Nevertheless, due to the nature of GAN-based models, they often suffer from unpleasant artifacts and unrealistic textures \cite{xie2023desra}. Besides, the adversarial training process is also unstable, which remains to be a challenging problem.
\subsection{Diffusion Probabilistic Models}
Inspired by non-equilibrium thermodynamics theory \cite{jarzynski1997equilibrium, neal2001annealed}, diffusion probabilistic models \cite{ho2020denoising} learn to generate data samples through a denoising sequence that estimate the score of the data distribution. DDPM \cite{ho2020denoising} is the trailblazer work that utilizes diffusion models for image generation with excellent experimental results that outperforms GAN-based methods. The main drawback of DDPM, however, is too computational expensive. It is not possible to apply DDPM on large images. DDIM \cite{song2020denoising} then proposed a different but efficient sampling method to partially solve the problem. The milestone work, LDM \cite{rombach2021highresolution}, applies diffusion model in the latent space of pre-trained autoencoders. It enables the training of large models with limited computational resources. Powerful text-to-image (T2I) models like Stable Diffusion based on latent diffusion have shown great generative ability in various downstream tasks \cite{ramesh2022hierarchical, saharia2022photorealistic, luo2021diffusion, 9775211} including image generation \cite{ho2022cascaded}, image editing \cite{mou2023t2i, zhang2023adding}, video generation \cite{wu2023tune, singer2022make}, etc.

\subsection{Diffusion Probabilistic Models for Image SR}
The first work to apply DDPM to image super-resolution is SR3 \cite{saharia2022image}. While this work is of great significance, its practical value is limited since it can only deal with simple degradation and relatively small image. Subsequently, researchers turn to pre-trained T2I models which have been proven that contains much stronger diffusion prior than GAN-based and other methods. ControlNet \cite{zhang2023adding} propose a novel way to input extra information like depth maps and edge maps in addition to text prompt. A number of works are then built on it. DiffBIR \cite{lin2023diffbir} adopts a two-stage pipeline and slightly adjusts ControlNet to fine-tune the pre-trained Stable Diffusion model. PASD \cite{yang2023pixel} proposes pixel-aware cross attention module to improve fidelity at pixel level. StableSR \cite{wang2023exploiting} employs time-aware encoder to harness to pre-trained model and controllable feature wrapping to balance the quality and fidelity of the generated image. However, since the information is severely damaged in the LQ image, they still fail to obtain high quality result with a lot of wrong content that do not belong to the original image. SeeSR \cite{wu2023seesr} and CoSeR \cite{sun2023coser} attempts to introduce more text information to the model. The former one trains a degradation-aware prompt extractor to generate soft and hard text prompt to provide additional representation information. The later one combines image appearance and language understanding to generate reference image, and proposes an All-in-Attention module to accept such reference. Yet additional information will inevitably result in additional computational cost, and it is not easy to generate text features that perfectly describe the complicated HR image. Hence, the goal of our approach is to efficiently generate additional information that can be used to activate the pre-trained model to obtain high-quality and high-fidelity results.

\section{Methodology}
\label{sec:method}
\subsection{Motivation and Overview}

\begin{figure}[tb]
  \centering
  \includegraphics[height=5.3cm]{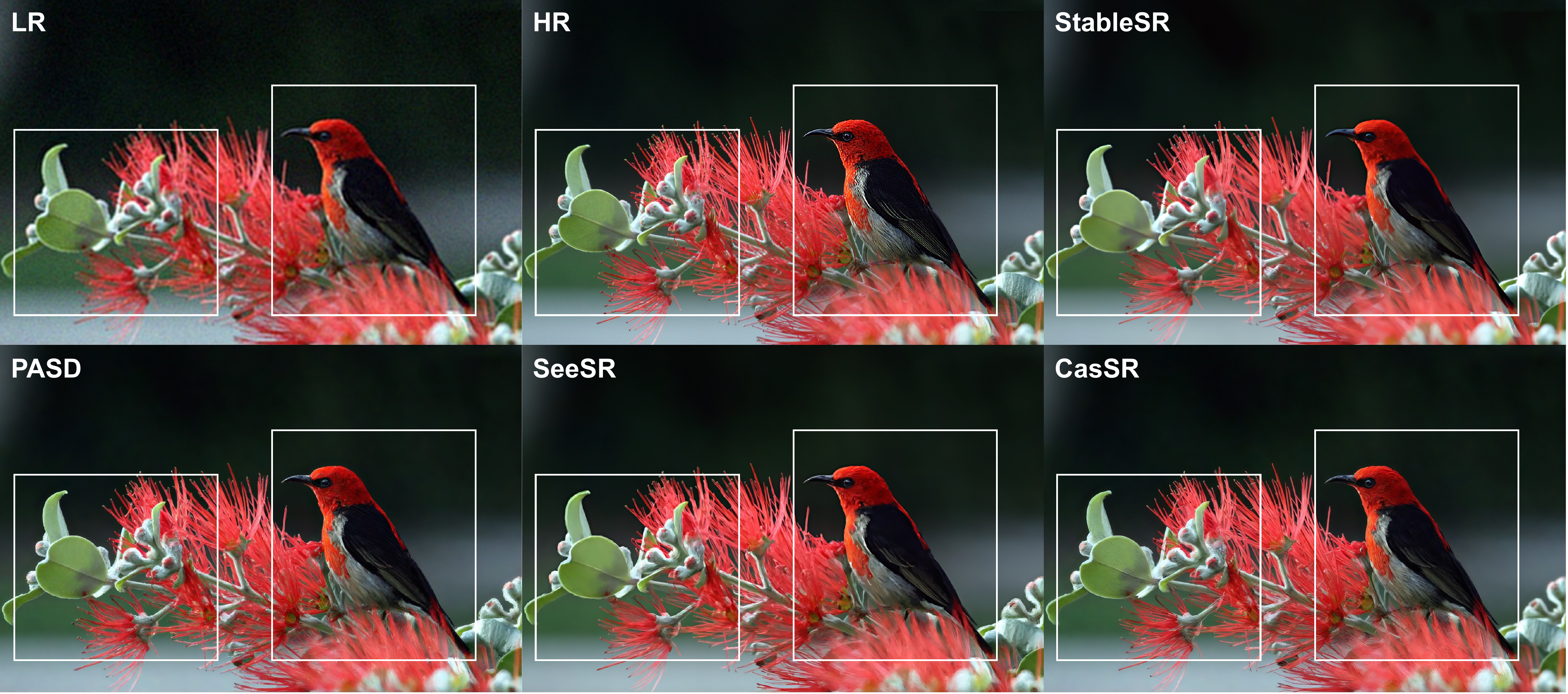}
  \caption{Comparison between our method and other state-of-the-art methods. Please pay attention to the highlighted areas. Our method accurately restored detailed textures of the objects. Zoom in for better view.}
  \vspace{-0.5cm}
  \label{fig:bird}
\end{figure}

\noindent\textbf{Motivation.} Stable-Diffusion based super-resolution methods has mainly focused on fully utilizing the prior knowledge in the pre-trained models. Models such as PASD \cite{yang2023pixel}, SeeSR \cite{wu2023seesr}, and CoSeR \cite{sun2023coser} have attempted to start from a linguistic perspective, guiding image generation through text prompt embeddings with more information. However, in real-world super-resolution problems, the input image is often complex. Hence, the commonly used classification style, tag style, and caption style text prompt are unable to provide a detailed description of the image.  As shown in Fig.~\ref{fig:bird}, Though PASD \cite{yang2023pixel} and SeeSR \cite{wu2023seesr} attempt to exploit semantic information, they still cannot correctly restore the detailed features like the feathers of birds and the patterns on leaves since it is almost impossible to describe them in just one or two sentences. In certain instances, as demonstrated in Sec~\ref{sec:ablation}, text prompts may even have a negative impact on the super-resolution results, since it can mislead Diffusion model and reduce the weight of the low-resolution image as an extra condition feature.

Moreover, we notice that the typical procedure in current image super-resolution algorithms involves firstly employing basic techniques like bicubic interpolation to enlarge the input low-resolution image to the intended resolution. The upsampled image is then utilized as the condition term for the T2I model to obtain the result. However, such a simple operation may add new disturbances to the image that is already heavily degraded, which cannot be compensated by adding text information. 

Based on the aforementioned observations, we posit that enhancing model performance through the lens of linguistic information presents a considerable challenge. When approaching the task from the perspective of textual prompts, the intricate task of crafting a text description that seamlessly aligns with image data proves to be difficult. Alternatively, delving into text embeddings introduces a layer of complexity that renders the process less transparent and intuitive. This motivate us to consider if more information could be activated from the input image itself to enhance the super-resolution results.

\begin{figure}[tb]
  \centering
  \includegraphics[height=5.4cm]{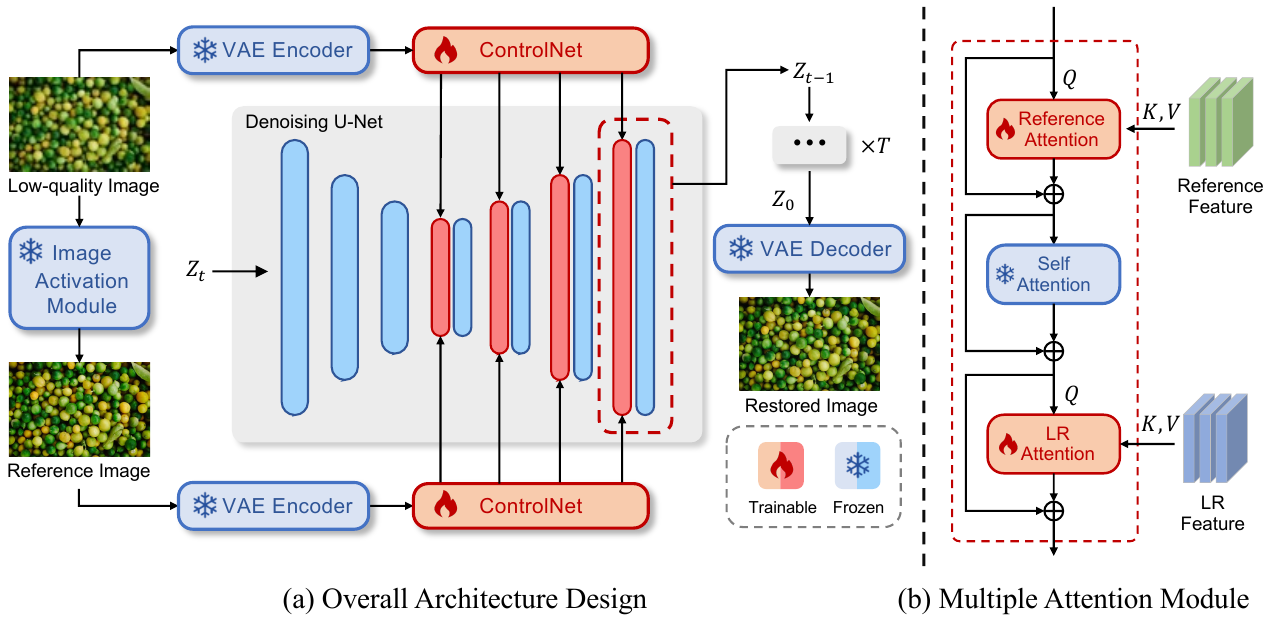}
  \caption{Architecture of the proposed Cascaded Super Resolution(CasSR) network.}
  \label{fig:archi}
\end{figure}

\noindent\textbf{Method Overview.}
We propose Cascaded Super Resolution (CasSR) to fully exploit the potential of the input image.  Specifically, the model contains two stages. In the first stage, an image activation is adopted to perform preliminary denoising and upsampling of the image and generate a reference image with more details and closer to the ground truth image. The generated image, together with the original low-quality (LQ) image, are sent to the second stage. The feature of the LQ image and the reference image are then extracted. We propose a multiple attention module to apply the above condition control to the pre-trained Stable Diffusion model. The architecture of the method is shown in Fig.~\ref{fig:archi}.

\subsection{Image Activation Module}
As previously discussed, we have introduced an image activation module aimed at optimizing images by a relatively straightforward denoising method, resulting in images with reduced degradation. In practice, we have found that the choice of model does not have a significant impact on the results, which is discussed in Sec~\ref{sec:ablation}. The model only needs to be able to initially remove the degradation of the input image. Specifically, we employ SCEdit \cite{jiang2023scedit}, a lightweight and computationally efficient controllable image diffusion generation model in our method. Intuitively, the generated image is expected to be of more detail and less degradation, and closer to the target HR image compared to the input image. This can provide more information to the pre-trained model to guide its diffusion process.

In comparison to CoSeR \cite{sun2023coser}, which also incorporates a reference image, it is worth noting that the reference image in CoSeR is generated through a cognitive embedding process involving both images and text. This approach is not straightforward and the generated images are typically new and different in comparison to the ground truth image, potentially introducing additional perturbations in certain scenarios.

\subsection{Multiple Image Guidance}
Next, we utilize the initial low-resolution image and the generated reference image as conditional inputs to the pre-trained Text-to-Image model. Like previous methods, we also adopt ControlNet \cite{zhang2023adding} as the controller for the T2I model. However, ControlNet is designed for high level control like edges, depth and pose, etc, and it may fail in fine super-resolution tasks. Hence, we introduce multiple attention module in the upsample blocks of the U-Net to enhance quality and fidelity of output by incorporating all conditional image's information.  

In particular, we adopt a weight-shared ControlNet to extract feature map from the low-resolutrion image and the reference image, denoted as $\mathbf{L}$ and $\mathbf{R}$, respectively. Initially, these feature maps are directly added to the feature extracted by the downsample module of the U-Net $\mathbf{X}$ as part of the conditional term. However, such a weak control makes it difficult to finely restore details in the image, especially in tasks like super-resolution that require pixel-level accuracy, which can lead to failure. They may generate inconsistent results, as shown in Fig~\ref{fig:ablation}.

To solve this problem, we utilize attention mechanism to enhance the control of conditional images. The query, key and value in attention mechanism are denoted as \textit{\textbf{Q}}, \textit{\textbf{K}} and \textit{\textbf{V}}, respectively. The reference feature $\mathbf{R}$ and the U-Net feature $\mathbf{X}$ are first fed into reference attention module, where $\bm{\mathit{Q_r}}$ is computed from $\mathbf{X}$, and $\bm{\mathit{K_r}}$, $\bm{\mathit{V_r}}$ is computed from $\mathbf{R}$. This block enables our model to generate higher quality image with better detail. The output, $\mathbf{X'}$, is computed as Eq.~\ref{eq:ca}
\begin{equation}
\label{eq:ca}
\mathbf{X'}=Softmax(\frac{\bm{\mathit{Q_r}}\bm{\mathit{K_r^T}}}{\sqrt{d}}) \cdot \bm{\mathit{V_r}}
\end{equation}
After performing the same self-attention operation on $\mathbf{X'}$ as in the oringinal U-Net model, $\mathbf{X'}$ and $\mathbf{L}$ are sent into LR attention module, where $\bm{\mathit{Q_{lr}}}$ is computed from $\mathbf{X'}$, and $\bm{\mathit{K_{lr}}}$, $\bm{\mathit{V_{lr}}}$ is computed from $\mathbf{L}$. Similarly, the output can be formulated as follow:
\begin{equation}
\mathbf{X_{out}}=Softmax(\frac{\bm{\mathit{Q_{lr}}}\bm{\mathit{K_{lr}^T}}}{\sqrt{d}}) \cdot \bm{\mathit{V_{lr}}}
\end{equation}
This is necessary because though reference image is of higher resolution, the simple super-resolution module may add artifacts to the image. Hence, the input low-resolution image is adopted to improve fidelity of the generated image. Subsequently, $\mathbf{X_{out}}$ is fed to next layer or the decoder for the final output.
\subsection{Training Objective}
\label{training}
In this section, we provide a detailed description of the training process for our model. First of all, we use a pre-trained SCEdit model as the image activation module to generate reference images. During the training phase, the ground truth image is encoded using a pre-trained Variational Autoencoder (VAE), with the resulting feature denoted as $z_0$. Subsequently, random noise is introduced to $z_0$, yielding $z_t$, where $t$ represents the diffusion step. The added noise is symbolized as $\epsilon$. By leveraging the features obtained from the low-resolution image $z_{lr}$ and the reference image $z_{r}$, the network $\epsilon_\theta$ estimates noise to recover $z_0$ from $z_t$. We find that language descriptions for images do not help in the denoising process, and may even have a negative impact. Therefore, during training and inference stages, we do not specifically design a text prompt, but only added some general descriptions such as “clean, high-resolution”, whose embedding denoted as $c$.The training objective, which is implemented by MSE Loss, can be expressed as Eq.~\ref{eq:loss}: 
\begin{equation}
\label{eq:loss}
\mathcal{L} = \mathbb{E}_{z_0,z_{lr},z_r,\epsilon \sim N }\lvert\lvert \epsilon - \epsilon_\theta(z_t,z_r,z_{lr},t,c)\rvert\rvert_2^2
\end{equation}
Throughout the training process, the parameters of the pre-trained Stable Diffusion model is frozen. All we need to train is the ControlNet model and the newly incorporated multiple cross-attention module within the U-Net architecture.
\section{Experiments}
\label{sec:exp}
\subsection{Experimental Settings} 
\noindent\textbf{Training dataset.}
Following previous works for fair comparison, We train CasSR on DIV2K \cite{agustsson2017ntire} , DIV8K \cite{gu2019div8k}, Flickr2K\cite{ timofte2017ntire }, OST \cite{wang2018recovering} and the first 10K images from FFHQ \cite{karras2019style}. We firstly random crop the HQ image to $512 \times 512$, and then use the degradation pipeline from Real-ESRGAN \cite{wang2021real} on the cropped HQ image to synthesize LR-HR image pairs.

\noindent\textbf{Testing dataset.}
To evaluate our method, we conduct experiments on both real-world dataset and synthesized dataset. For real-world dataset, we employ two bench mark dataset, DRealSR \cite{wei2020component} and RealSR \cite{cai2019toward}. To obtain LR-HR image pairs, we adopt the same cropping method as \cite{wang2023exploiting}. Specifically, We center crop the LR image to $128 \times 128$, and the HR image to $512 \times 512$. For synthesized dataset, we employ the validation set of DIV2K \cite{agustsson2017ntire} dataset. It is worth noting that we simply apply the degradation pipeline from Real-ESRGAN \cite{wang2021real} on the original HQ image, and we do not perform any cropping on the original image, which means that the resulting LQ-HQ image pair is relatively large.

\noindent\textbf{Implementation Details.} We employ SD-1.5 \cite{rombach2021highresolution} as the pre-trained model, and adopt Adam \cite{kingma2014adam} optimizer to train the model with a batch size of 4. The learning rate is set as $5\times 10^{-4}$. Both models, including SCEdit \cite{jiang2023scedit} used in image activation module and the super resolution module in the second stage, are trained on 8 Tesla-A100 GPUs with the same settings. It is also fine to simply adopt a pre-trained image super-resolution model as image activation module to reduce training time, which is discussed in \ref{sec:ablation}. According to our experiments, 100K steps is enough for the training, which costs about 2 days.

During inference, we integrate the LR image into the initial noise. Previous researches \cite{wu2023seesr, yang2023pixel} prove that it can provide a better start point for the diffusion process. For each input, we use the generic description same as described in Sec~\ref{training}. In addition, we also utilize Classifier-Free Guidance(CFG) \cite{ho2022classifier} technique, which have been proven to be very effective, to generate high-quality image.

\noindent\textbf{Compare Methods.} We compare CasSR with state-of-the-art real-world super-resolution methods, including both GAN-based methods like  BSRGAN \cite{zhang2021designing}, FeMaSR \cite{chen2022real}, HAT \cite{chen2023activating} and DASR \cite{wei2021unsupervised}, and diffusion-based methods like PASD \cite{yang2023pixel}, StableSR \cite{wang2023exploiting}, DiffBIR \cite{lin2023diffbir} and SeeSR \cite{wu2023seesr}. We use the publicy released code and trained models to conduct the experiments.

\noindent\textbf{Evaluation Metrics.}For quantitative evaluation, we employ numerous widely used metrics, including FID \cite{heusel2017gans}, LPIPS \cite{zhang2018unreasonable}, DISTS \cite{ding2020image}, MUSIQ \cite{ke2021musiq}, MANIQA \cite{yang2022maniqa}, CLIPIQA \cite{wang2022exploring} and NIQE \cite{mittal2012making}. It is worth noting that we do not adopt PSNR and SSIM \cite{wang2004image} which are pixel-level metrics, which has already been discussed by previous researches \cite{yu2024scaling, sun2023coser}. The reason is that they can not completely reflect the quality of the image as perceived by humans. LPIPS \cite{zhang2018unreasonable}, DISTS \cite{ding2020image} and FID \cite{heusel2017gans} are reference-based perceptual metrics, measuring the distance and similarity between the ground truth image and SR image. NIQE \cite{mittal2012making}, MUSIQ \cite{ke2021musiq}, CLIPIQA \cite{wang2022exploring} and MANIQA \cite{yang2022maniqa} are non-reference perceptual metrics which assess the quality of the generated image. We employ the IQA-Pytorch \cite{pyiqa} to compute these metrics.

\begin{figure}[tb]
  \centering
  \includegraphics[height=7.0cm]{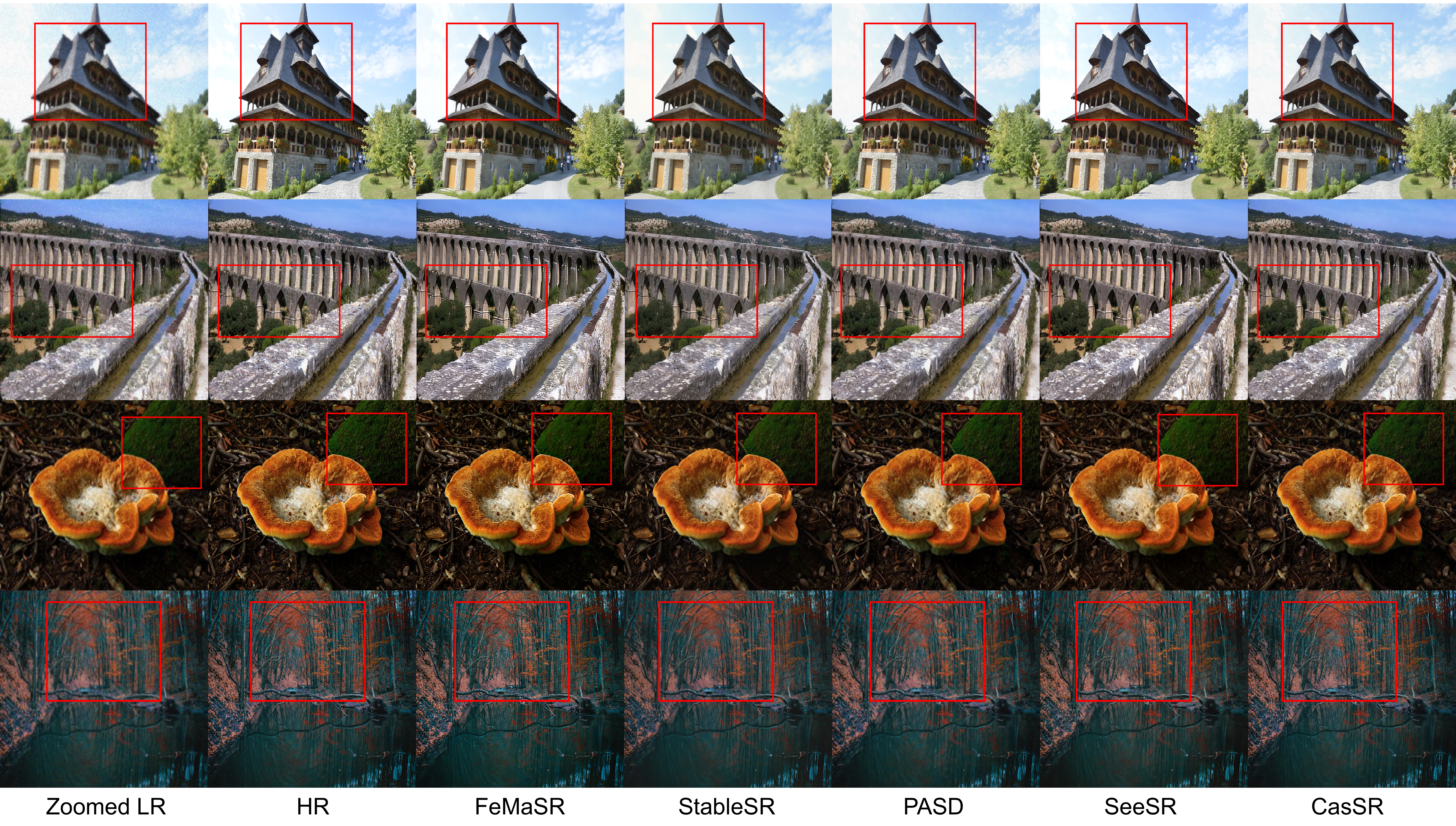}
  \caption{Qualitative comparisons of different methods on real image super-resolution. Please focus on the areas marked by red boxes, and zoom in for a better view.}
  \label{fig:qualitative}
\end{figure}
\subsection{Experimental Results}
\textbf{Quantitative Comparison.}
The quantitative results presented in Table~\ref{table:quantitative} showcase the performance of our method across 3 real-world and synthetic datasets. Our experimental findings consistently highlight the exceptional performance of our approach, particularly excelling in perceptual metrics that assess image quality. Notably, when evaluated on the DIV2K \cite{agustsson2017ntire} validation dataset, known for its larger images, our model consistently achieves either the best or second best results across all metrics, establishing it as a valuable asset for practical applications. Specifically, our MUSIQ score outperforms the second best method by 3.82\%, while our MANIQA score surpasses the second best method by 7.05\%. Furthermore, in the DRealSR \cite{wei2020component} and RealSR \cite{cai2019toward} datasets, our method demonstrates competitive performance. We argue that the subpar performance on some metrics is due to the fact that the input images are cropped, resulting in information loss which has a bigger impact on our method than others. Nevertheless, the results from non-reference metrics affirm the quality of the images generated by our method.

\noindent\textbf{Qualitative Comparison.}
The general visual comparison is shown in Fig~\ref{fig:qualitative}. Since our model fully activates the diffusion prior and enriches the input information, it produces results with good fidelity, especially in intricate details where our generated results are closer to the original image. The image super-resolution algorithm based on the pre-trained Diffusion model is prone to generating artifacts that do not belong to the original image, and experimental results show that our method can also avoid this issue. 

\begin{figure}[tb]
  \centering
  \includegraphics[height=4.0cm]{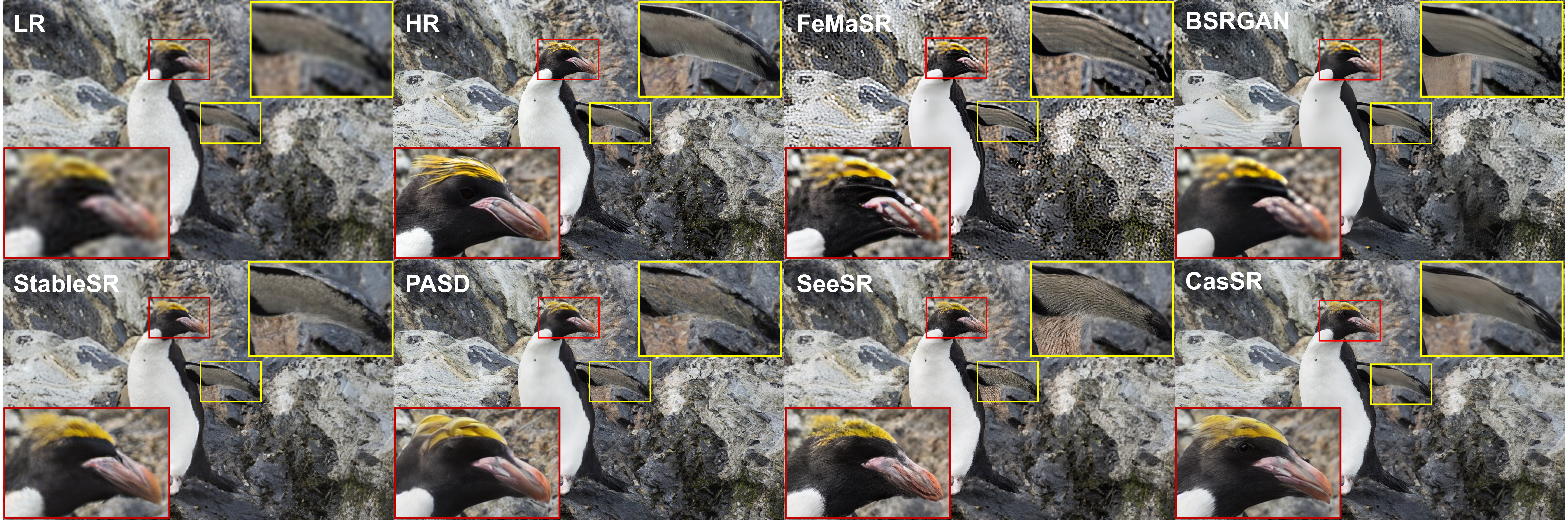}
  \caption{Qualitative comparison on LQ image. CasSR successfully recovers the penguin's eye and also avoids confusing its wings with background rocks. Comparing to other methods, CasSR output preserves better details existing in the LQ image.}
  \label{fig:penguin}
\end{figure}

According to the quantitative experimental results, we select FeMaSR \cite{chen2022real} and BSRGAN \cite{zhang2021designing} as the representative GAN-based method. According to Fig~\ref{fig:qualitative} and Fig~\ref{fig:penguin}, it is obvious that they cannot generate high quality results when the input image is relatively large and degradation is severe, though they may have good quantitative results.

As for diffusion-based methods shown in Fig~\ref{fig:qualitative}, specifically, as shown in the first row of Fig~\ref{fig:qualitative}, other methods cannot generate the texture details on the roof well. This is because the input image is too blurry, even though StableSR \cite{wang2023exploiting} and PASD \cite{yang2023pixel} have noticed this and design modules to enhance quality based on the input image, they still cannot obtain enough reference information. A similar situation can be seen in the second row, where the clarity of the bushes and the patterns on the bridge stones show that the images produced by our method are of the highest quality. The image in the third row demonstrates the advantage of our method in fidelity. Notice the moss in the top right corner, only our method accurately restored it. This is because our model initially enhances the input image, activates more information, and therefore has stronger guidance for generating images. Last but not the least, the fourth row overall showcases the generation quality of our method. A more detailed demonstration of our super-resolution result can be found in Fig~\ref{fig:penguin}. All other methods fail to correctly restore the penguin's wing and confuse it with the background rocks. Moreover, they also cannot find the penguin's eye which has the same color with its head. However, our method is able to address this issue and generate a image of high fidelity.

\begin{table*}
\caption{Quantitative comparisons on both synthetic and real-world benchmarks. The best and second best are highlighted in \textcolor{red}{red} and \textcolor{blue}{blue}. All methods are tested with their officially released trained models.}
\begin{center}
\resizebox{\textwidth}{!}{
\begin{tabular}{c|c|cccc|cccc|c}
\toprule

Dataset & Metrics &
 BSRGAN \cite{zhang2021designing}  & FeMaSR \cite{chen2022real}& HAT \cite{chen2023activating} & DASR \cite{wei2021unsupervised}  &
 PASD \cite{yang2023pixel} & StableSR \cite{wang2023exploiting} & DiffBIR \cite{lin2023diffbir}& SeeSR \cite{wu2023seesr}& \textbf{CasSR(Ours)}  \\
\midrule

\multirow{7}{*}{DIV2K-val \cite{agustsson2017ntire}} 
    
    & FID $\downarrow$  &
        52.9180 & 42.2293 & 41.7034 & 55.3689 & 
        31.6253 & 31.2679 & 39.4041 & \textcolor{red}{28.5568} &\textcolor{blue} {30.7808}
        \\
    & DISTS $\downarrow$ &
        0.1899  & 0.1617& 0.1919 & 0.3068 & 
         0.1599 &  0.1596 & \textcolor{red}{0.1558} & 0.1601 & \textcolor{blue}{0.1531}
        \\
    & LPIPS $\downarrow$   &
        0.3516  & 0.3272 & 0.3175 & 0.4505 & 
        0.3254 & 0.3316 & 0.3658 & \textcolor{red}{0.3050} &\textcolor{blue}{0.3225}
        \\
    & MUSIQ $\uparrow$   &
        62.5478  & 60.8010 & 57.0293 & 53.0613 & 
        65.7858 & 55.9177 & \textcolor{blue}{67.0657} & 66.3587 & \textcolor{red}{68.6242}
        \\
    & MANIQA $\uparrow$   &
        0.3489  & 0.3334 & 0.3062 & 0.2933 & 
        0.3920 & 0.2958 & \textcolor{blue}{0.4580}& 0.4393 & \textcolor{red}{0.4927}
        \\
    & CLIPIQA $\uparrow$   &
        0.5623  & 0.6167 & 0.4486 & 0.4603 & 
        0.5933 & 0.5053 & \textcolor{red}{0.7018} & 0.6277 & \textcolor{blue}{0.6861}
        \\
    & NIQE $\downarrow$   &
        3.7414  & 3.8060 & 4.1641 & 4.5046 & 
        3.6746 & 4.5062 & \textcolor{red}{2.9574} & 3.8750 & \textcolor{blue}{3.6026}
        \\

\midrule

\multirow{7}{*}{DRealSR \cite{wei2020component}} 
    %
    & FID $\downarrow$  &
        157.0340  & 157.7239 & \textcolor{red}{141.8644} & 170.1904 & 
        151.7256 & 145.6182 & 170.5032 & \textcolor{blue}{145.1622} & 166.1467
        \\
    & DISTS $\downarrow$ &
        0.2352  & \textcolor{blue}{0.2270} & 0.2462 & 0.2697 & 
         0.2544 & \textcolor{red}{0.2073} & 0.2683 & 0.2342 & 0.2868 
        \\
    & LPIPS $\downarrow$   &
        0.3600 & 0.3255 & 0.3370 & 0.3705 & 
        0.3776 & \textcolor{red}{0.2779} & 0.4447 & 0.3301 & \textcolor{blue}{0.3067} 
        \\
    & MUSIQ $\uparrow$   &
        57.1217  & 57.1217 & 50.3458 & 40.6016 & 
        54.9425 & 51.5160 & 60.8363 & \textcolor{blue}{64.7460} & \textcolor{red}{67.3450} 
        \\
    & MANIQA $\uparrow$   &
        0.3420  & 0.3155 & 0.3102 & 0.2700 & 
        0.3917 & 0.3235 & {0.4486} & \textcolor{red}{0.5048} & \textcolor{blue}{0.4985}
        \\
    & CLIPIQA $\uparrow$   &
        0.5084  &  0.5625& 0.3901 & 0.4700 & 
        0.5478 & 0.5126 & 0.6476 & \textcolor{blue}{0.6883} & \textcolor{red}{0.7142}
        \\
    & NIQE $\downarrow$   &
        6.5994 & \textcolor{blue}{5.9545} & 7.2654 & 8.2681 & 
        7.0751 & 7.5105 & 6.1496 & 6.3751 &\textcolor{red}{5.6890} 
        \\
\midrule

\multirow{7}{*}{RealSR \cite{cai2019toward}} 
    & FID $\downarrow$  &
        143.3876 & 140.2071 & \textcolor{red}{123.0671} & 163.8753 & 
        \textcolor{blue}{126.6657} & 129.3065 & 130.4431 & 126.9731 & 139.0698
        \\
    & DISTS $\downarrow$ &
        0.2125  & 0.2263 & \textcolor{red}{0.1996} & 0.2662 & 
         0.2066 &  \textcolor{blue}{0.2034} & 0.2313 & 0.2264 & 0.2459
        \\
    & LPIPS $\downarrow$   &
        \textcolor{blue}{0.2688}  & 0.2918 & 0.2469 & 0.3585 & 
        0.2788 & \textcolor{red}{0.2657} & 0.3644 & 0.3034 & 0.3822
        \\
    & MUSIQ $\uparrow$   &
        63.2816  & 58.5714 & 57.3266 & 46.8467 & 
        62.9266 & 61.0109 & {64.6228} & \textcolor{blue}{69.7639} & \textcolor{red}{70.4224}
        \\
    & MANIQA $\uparrow$   &
        0.3782  & 0.3510 & 0.3364 & 0.2585 & 
        0.4080 & 0.3667 & {0.4545}& \textcolor{red}{0.5463} & \textcolor{blue}{0.5326}
        \\
    & CLIPIQA $\uparrow$   &
        0.5076  & 0.5353 & 0.3793 & 0.3844 & 
        0.5190 & 0.5218 & {0.6521} & \textcolor{blue}{0.6795} &\textcolor{red}{0.6968}
        \\
    & NIQE $\downarrow$   &
        5.7083  & 5.6371 & 6.4163 & 6.9204 & 
        5.6479 & 6.5735 & 5.7470 & \textcolor{red}{5.2362} & \textcolor{blue}{5.3491}
        \\

\bottomrule

\end{tabular}
}
\end{center}
\label{table:quantitative}
\end{table*}
\subsection{Ablation Study}
\label{sec:ablation}
\begin{figure}[tb]
  \centering
  \includegraphics[height=3.58cm]{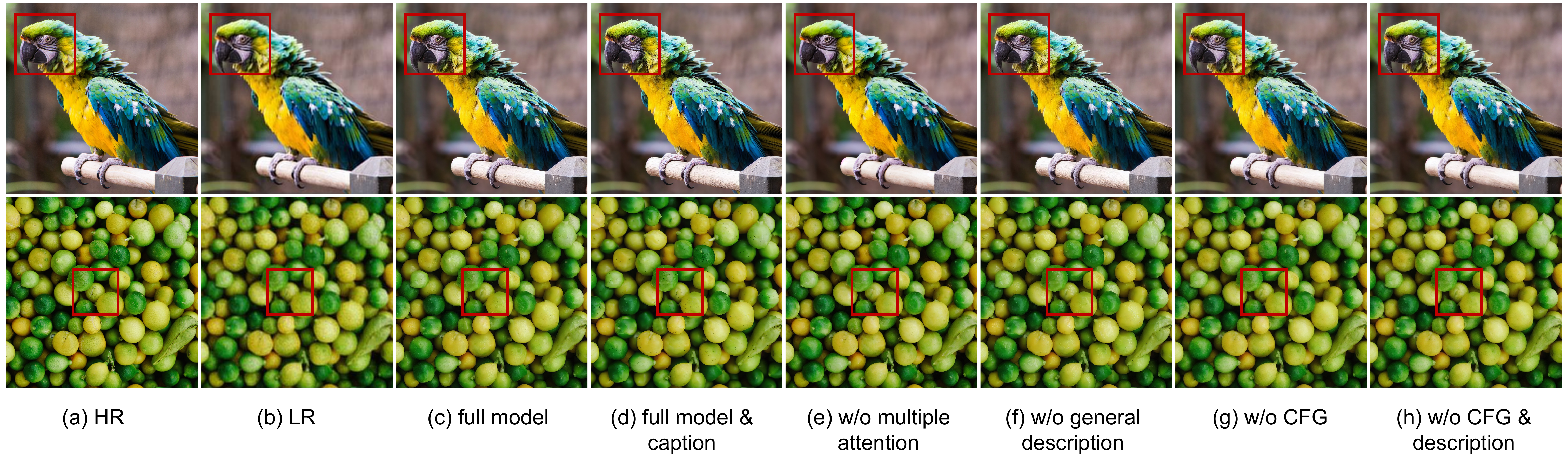}
  \caption{Ablation study on different variants of CasSR. Zoom in for a better view.}
  \label{fig:ablation}
\end{figure}

\textbf{Importance of multiple attention module.} We first show the importance of multiple attention module by excluding it from the U-Net, {\it i.e.}, the reference image feature $\mathbf{R}$ and the LR image feature $\mathbf{L}$ are directly added to the U-Net feature $\mathbf{X}$ as additional residuals. Actually, at this point our method has degenerated into a method similar to ControlNet \cite{zhang2023adding}. As shown in Fig~\ref{fig:ablation}(e), the output quality has decreased with much less details comparing to other results, highlighting the importance of the multiple attention module.

\noindent\textbf{Importance of image activation module.} To delve deeper into the impact of the reference image, we explored various approaches, such as utilizing the LQ image directly as the reference image and generating the reference image through methods like ControlNet \cite{zhang2023adding} and SCEdit \cite{jiang2023scedit}. As shown in Table~\ref{table:ablation}, incorporating the reference image can significantly improve the results. However, the method used in the image activation module to generate the reference image does not have strict requirements. It is worth noting that these methods are not tailored specifically for image super-resolution, hence the results may not be particularly refined. This also underscores the robustness of our model.
\begin{table}
\vspace{-0.3cm}
\caption{Ablation study on the importance of image activation module and the choice of corresponding model. The best result is highlighted in \textcolor{red}{red}.}  
\vspace{-0.3cm}
\begin{center}
\resizebox{\textwidth}{!}{
\scriptsize
\begin{tabular}{c|ccccc}
\toprule

 Reference & FID $\downarrow$ & DISTS $\downarrow$ & LPIPS  $\downarrow$ & MUSIQ $\uparrow$ & MANIQA $\uparrow$ \\
\midrule    
    LR-Image  &
        33.1480 & 0.1797 & 0.3216 & 63.4204 &  0.4534 
        \\
    ControlNet \cite{zhang2023adding} &
        \textcolor{red}{30.5114}  & 0.1534& 0.3207 & \textcolor{red}{69.0435} &  \textcolor{red}{0.5037} 
        \\
    SCEdit \cite{jiang2023scedit}   &
        30.7808  & \textcolor{red}{0.1531} & \textcolor{red}{0.3175} & 68.6242 & 0.4927
        \\
\bottomrule
\end{tabular}

}
\end{center}  
\vspace{-0.8cm}
\label{table:ablation}
\end{table}

\noindent\textbf{Effect of semantic feature.} We experimented with various text prompt designs, such as with or without Classifier-Free Guidance (CFG) \cite{ho2022classifier}, null inputs, and general descriptions (such as “clean, high-resolution”). Additionally, we explored designs akin to the high-level information proposed in PASD \cite{yang2023pixel} which integrates text prompts derived from captioning. As shown in Table~\ref{table:text}, relatively general methods had a greater impact on the model's performance, while finely crafted text prompts did not yield the outstanding results. 
\begin{table}
\vspace{-0.3cm}
\caption{Ablation study on the effect of semantic feature. The
best result is highlighted in \textcolor{red}{red}.}
\vspace{-0.3cm}
\begin{center}
\resizebox{0.99\textwidth}{!}{
\begin{tabular}{ccc|ccccc}
\toprule
General Description & CFG & Caption &  FID $\downarrow$ & DISTS $\downarrow$ & LPIPS  $\downarrow$ & MUSIQ $\uparrow$ & MANIQA $\uparrow$ \\
\midrule    
     $\checkmark$ &  $\checkmark$  & $\times$  &
        30.7808 & \textcolor{red}{0.1531} & \textcolor{red}{0.3225} & \textcolor{red}{68.6242} &  \textcolor{red}{0.4927} 
        \\
    $\times$ &  $\times$  & $\times$  &
        32.6440  & 0.1844 & 0.3141 & 57.3091 &  0.3260 
        \\
    $\times$ &  $\checkmark$  & $\times$  &
        32.4908  & 0.1572 & 0.3250 & 68.5192 &  0.4773 
        \\
    $\checkmark$ &  $\checkmark$  & $\checkmark$   &
        \textcolor{red}{30.6618}  & 0.1535 & 0.3280 & 68.2996 & 0.5057
        \\
\bottomrule
\vspace{-1.0cm}

\end{tabular}

}
\end{center}
\vspace{-0.3cm}
\label{table:text}
\end{table}

We show a qualitative comparison of ablation study in Fig~\ref{fig:ablation}. As shown in the figure, adding general description and CFG can both make the image much clear. In contrast, using a specific caption tailored for a particular image may not yield consistently good results. In Fig~\ref{fig:ablation}(d), we can clearly find unpleasant feather around the bird's eye and black dots on the fruit, which do not exist in the HR image. This observation validates our argument, {\it i.e.}, Compared to focusing on designing better text prompts and semantic embeddings, it is easier to improve the SR results by trying to explore more the information in images that has not been fully utilized.

\section{Conclusion}
We propose CasSR, a stable-diffusion based super resolution method utilizing image information to enhance the generation capability. With a simple but effective multiple attention module, the generated reference image together with the original input image can guide the model to generate high quality and fidelity results. Numerous experiments demonstrate the exceptional efficacy of our approach in terms of quantitative and qualitative metrics, particularly on datasets characterized by large-scale and intricate imagery. This underscores the practical utility of our model.

%
%
\bibliographystyle{splncs04}
\bibliography{egbib}
\end{document}